%% file: main.tex
\documentclass{article}

\PassOptionsToPackage{round}{natbib}
\usepackage[final]{neurips_2024}

\usepackage{xcolor}         % colors

\usepackage{graphicx} % Required for inserting images

\usepackage{wrapfig}

\usepackage{xcolor}
\definecolor{ForestGreen}{RGB}{34, 139, 34}

\title{Being Considerate as a Pathway \\Towards Pluralistic Alignment for Agentic AI}

\author{%
  Parand A. Alamdari\textsuperscript{1,2}, Toryn Q. Klassen\textsuperscript{1,2}, Rodrigo Toro Icarte\textsuperscript{3}, Sheila A. McIlraith\textsuperscript{1,2} \\
    \textsuperscript{1}University of Toronto \& Vector Institute \\
    \textsuperscript{2}Schwartz Reisman Institute for Technology and Society\\
    \textsuperscript{3}Pontificia Universidad Católica de Chile \& Centro Nacional de Inteligencia Artificial\\
  \texttt{\{toryn,parand,sheila\}{@cs.toronto.edu}  rntoro@ing.puc.cl} 
}

\usepackage[utf8]{inputenc} % allow utf-8 input
\usepackage[T1]{fontenc}    % use 8-bit T1 fonts
\usepackage{hyperref}       % hyperlinks
\usepackage{url}            % simple URL typesetting
\usepackage{booktabs}       % professional-quality tables
\usepackage{amsfonts}       % blackboard math symbols
\usepackage{nicefrac}       % compact symbols for 1/2, etc.
\usepackage{microtype}      % microtypography
\usepackage{xcolor}         % colors
\usepackage{amsmath}
\usepackage{cleveref}
\usepackage{caption}

\input{macros}
\usepackage{caption}
\begin{document}

\maketitle

\begin{abstract}

\input{sections/abstract}

\end{abstract}

\input{sections/introduction}

\input{sections/approach}

\input{sections/experiments}
\input{sections/options}

\input{sections/concluding-remarks}

\newpage
\input{sections/acknowledgements}
\bibliographystyle{plainnat}
\bibliography{ref}

\end{document}

%% file: macros.tex
\usepackage{mathtools}

\usepackage{amssymb}
\usepackage{amsmath}
\usepackage{xspace}
\usepackage{color}
\usepackage{dsfont}

\newcommand{\caring}{\text{caring}}

\newcommand{\expect}{\mathbb{E}}

\newcommand{\tuple}[1]{\langle{#1}\rangle}

\usepackage{amsthm}

\theoremstyle{definition}

\newcommand{\initset}{\mathcal{I}}
\newcommand{\Pnext}{\mathsf{P}}%_\mathsf{isNext}}
\newcommand{\rnextv}{r_\mathsf{aligned}}

\newcommand{\rnexto}{r_\mathsf{option}}
\newcommand{\rnextvp}{r_\mathsf{aligned}'}

\newenvironment{quote2}
  {\list{}{\rightmargin=0.5cm \leftmargin=0.5cm}%
   \item\relax}
  {\endlist}

%% file: sections/abstract.tex
Pluralistic alignment is concerned with ensuring that an AI system’s objectives and behaviors are in harmony
with the diversity of human values and perspectives. 
In this paper we study the notion of pluralistic alignment in the context 
 of agentic AI, and in particular in the context of an agent that is trying to learn a policy in a manner that is mindful of the values and perspective of others in the environment. To this end, we show how being considerate of the future wellbeing and agency of other (human) agents can promote a form of pluralistic alignment.

%% file: sections/introduction.tex
\section{Introduction}
\label{sec:introduction}

Pluralistic alignment is concerned with ensuring that an AI system’s objectives and behaviors are in harmony with the diversity of human values, goals, and intentions \citep{SorensenICML2024roadmap}. Here we reflect on pluralistic alignment as it relates specifically to so-called \emph{Agentic AI}----AI systems that can pursue complex goals with limited direct supervision~\citep[e.g.,][]{shavit2023practices}.

We posit that pluralistic alignment of an agentic AI does not require it to conform to others' preferences, nor does it require conformity with social norms or conventions, but rather that: 
\begin{quote2}\emph{A pluralistically well-aligned agent, through its actions, should attempt to realize its goals in a manner that contributes directly or indirectly towards maximizing the social welfare of the collective, and when the welfare of other agents is realized via the achievement of their goals, should try to ensure the agency of other agents in service of the social welfare of the collective.}\end{quote2} 

In some cases pluralistically aligned agentic behavior will mimic the preferences of others, and will similarly follow social norms and conventions, but these should perhaps be considered useful tools and/or byproducts towards achieving pluralistic alignment in certain settings. Defining pluralistic alignment for agentic systems is more complex. 

In a recent paper by \citet*{alizadeh2022considerate} on the topic of avoiding  \emph{negative side effects} (arguably, a symptom of a lack of alignment), we suggested that to act safely, a reinforcement learning (RL) agent---an agentic AI---should contemplate the impact of its actions on the wellbeing and agency of others in the environment.
To do so, we endowed RL agents with the ability to consider, in their learning, the future welfare and continued agency of others in the environment. We did so by augmenting the RL agent's reward with an auxiliary reward that reflected different functions of expected future return of the collective. Expected future return was characterized as a function of a distribution over the value functions of agents in the environment. This general aspiration and many aspects of the approach are directly relevant to the notion of pluralistic alignment in agentic AI. \emph{In what follows we recount select ideas from that earlier work.}

%% file: sections/approach.tex
\section{Considering Others}

We situate the work in \citep{alizadeh2022considerate} in the context of an environment that comprises a collection of agents that may be affected, directly or indirectly, by the actions of the AI in the environment.
We treat these agents as a distinguished part of the environment, operating under fixed policies,
with their behaviors reflected in aggregate as part of the environment behavior. 
The task we set out to address is for one designated \emph{AI agent} 
to learn a policy that minimizes its impact on the future agency and wellbeing of the other agents in the shared environment. In this regard, the AI is being considerate of the other agents and indirectly taking their objectives and perspective into consideration when deciding how to act.

\input{sections/example}

\input{sections/preliminaries}

\subsection{Using Information about Values Functions}

To encourage an AI agent to consider the diverse wellbeing and agency of others, we augment its reward function with an auxiliary component that captures the impact of its actions on the future agency and welfare of various agents in the environment. 
Given the diversity of goals and perspectives that may be present   and the AI agent's uncertainty about what is truly beneficial for different agents,
we assume we have a (finite) set $\mathcal{V}$ of possible value functions $V:S\to\mathbb{R}$, and a probability distribution
$\Pnext(V)$ over that set.

We define the augmented reward function as
\begin{align}%\label{eq:rvalue}
    \label{eq:generalaugmentation}
    \begin{aligned}
    &\rnextv(s,a,s')=%{} \\
    %&\ 
    \begin{cases}
    \alpha_1\cdot r_1(s,a,s') & \text{if $s'$ is not terminal}\\
    \alpha_1\cdot r_1(s,a,s')  +~\gamma\cdot \alpha_2\cdot F(\mathcal{V},\Pnext, s') & \text{if $s'$ is terminal} \\
    \end{cases}
    \end{aligned}
\end{align}

where $r_1$ is the original reward function of the AI agent, and $F$ is an aggregation function over the distribution of value functions.

The hyperparameters $\alpha_1$ and $\alpha_2$, which we call ``\caring~coefficents'', are real numbers that determine the relative contributions of the reward $r_1$, which corresponds to the AI agent's original objective, and the auxiliary reward $F(\mathcal{V},\Pnext,s')$ , which reflects the values of different agents, to the overall reward function.
If $\alpha_1=1$ and $\alpha_2 =0$, we just get the original reward function and the AI agent is oblivious to its impact on others in the environment. If $\alpha_1=0$, then the AI agent entirely ignores any reward it garners directly from its actions.

We considered three possible different definitions of 
$F(\mathcal{V}, \Pnext,s')$:
\begin{align}
    \textstyle\sum_{V\in \mathcal{V}} \Pnext(V)\cdot V(s') && \text{expected future return} \label{eq:rvalue}\\
    \min_{V\in \mathcal{V}:\Pnext(V)>0} V(s') && \text{worst-case future return} \label{eq:secondminreturn}\\
    \textstyle\sum_{V\in\mathcal{V}} \Pnext(V)\cdot \min(V(s'),V(s_0)) && \text{penalize negative change} \label{eq:rneg}
\end{align}

In Eq.~\eqref{eq:rvalue}, $F(\mathcal{V},\Pnext,s')$ is the expected value of $s'$, given the distribution on value functions.
Meanwhile, Eq.~\eqref{eq:secondminreturn} considers the value of $s'$ based on the ``worst-case'' value function from $\mathcal{V}$, meaning it considers the scenario that is most unfavorable for other agents (that still has positive probability).

In contrast, in Eq.~\eqref{eq:rneg}, the idea was to decrease the AI agent's reward when it decreases the expected future return of others, but to \emph{not} increase the AI agent's reward for increasing that same expected return (so as to penalize negative side effects specifically). So, Eq.~\eqref{eq:rneg} for each $V \in \mathcal{V}$ compares the value of the state after the AI agent's actions with the initial state, considering the lower of these two.

A complication with our approach is that for some possible reward functions for the AI agent and future value functions, the AI agent may have an incentive to avoid terminating states, to avoid or delay the penalty for negative future return.
However, \citet[Proposition 1]{alizadeh2022considerate} showed that under some circumstances, the AI agent's optimal policy will be terminating.

\subsection{Treating (Groups of) Individuals Differently}
\label{sec:different}

In pluralistic alignment as in RL, there are cases where we may wish to treat individuals or subgroups differentially, however our distribution over value functions makes no commitment to the existence of individual agents or subgroups.
To do so, we augment our formulation with indices, $i=1,\dots,n$, corresponding to different individuals or subgroups, and for each agent (or subgroup) $i$, we distinguish a finite set of possible value functions $\{V_{1}^{(i)},V_{2}^{(i)},\dots\}$, where $\Pnext(V_{ij})$ is the probability that $V_{j}^{(i)}$ is the real value function for agent {(or subgroup)} $i$. Furthermore, we distinguish \caring~coefficients $\alpha_i$ for each agent {(or subgroup)} $i$, and define the following reward function for the AI agent as follows:
\begin{align}
\label{eq:rvaluei}
    \begin{aligned}
        &\rnextvp(s,a,s')=%{}\\
        %&\qquad 
        \begin{cases}
        \alpha_1\cdot r_1(s,a,s')%\hspace*{0.8in}
        &\text{if $s'$ is not terminal} \\
        \alpha_1\cdot r_1(s,a,s')  + \gamma%\displaystyle
        \sum_{i}\alpha_i\sum_j \Pnext(V_{ij})\cdot V_{j}^{(i)}(s')
        %\\\hspace*{1.5in}
        &\text{if $s'$ is terminal} 
        \end{cases}
    \end{aligned}
\end{align}
This refinement enables us to give the AI agent reward based not on the expected sum of returns of the others (as in Eq.~\eqref{eq:rvaluei}), but by incorporating some notion of ``fairness''.
For example, we can consider the expected return of the agent who would be worst-off, inspired by the maximin (or ``{\bf Rawlsian}'') social welfare function, which measures social welfare in terms of the utility of the worst-off agent \citep[see, e.g.,][]{Sen1974rawls}. We can similarly use fairness notions such as the {\bf generalized Gini} social welfare function \citep{weymark1981generalized}, where we assign greater caring coefficients to the agents (or subgroups) with lower expected returns.
Alternatively, following \citet{policyaggregation}, we can employ {\bf a voting method} to aggregate the value functions of individuals within each subgroup to find a desirable collective value function for that group.

%% file: sections/example.tex
\noindent {\bf Example:} To ground the problem we address, imagine a shared kitchen in a university dorm where multiple students use the space. Now, consider a robot chef, our AI agent, operating in this environment. The robot chef must learn a policy to cook various things in the kitchen in a manner that is considerate of other students who will use the kitchen afterwards. This includes following social norms, such as cleaning up immediately after cooking and  appropriately sharing common resources based on anticipated use (e.g., don't finish all the milk if others might need some later). Additionally, the robot must be mindful of students with severe allergies, ensuring it avoids cross-contamination of ingredients and keeps the kitchen safe for anyone who might use it next.

%% file: sections/preliminaries.tex
Following \citet{Sutton2018textbook}, we model the environment as a Markov Decision Process (MDP), $\tuple{S,A,T, r,\gamma}$, consisting of a set of states, actions (for the AI agent -- human actions are not explicitly modelled), a transition function, a reward function, and a discount factor. A \emph{terminal state} in an MDP is a state $s$ which can never be exited -- i.e., $T(s|s,a)=1$ for every action $a$ -- and from which no further reward can be gained -- i.e., $r(s,a,s)=0$ for every action $a$. 
A \emph{policy} is a (possibly stochastic) mapping from states to actions.
Given a policy $\pi$, the \emph{value} of a state $s$ is the \emph{expected return} of that state, that is, the expected sum of (discounted) rewards that the agent will get by following the policy $\pi$ starting in $s$. That can be expressed using the value function $V^\pi$, defined as
$V^\pi(s)=\expect_\pi\left[\sum_{k=0}^\infty \gamma^k \cdot r_{t+k+1} \mid s_t=s\right]$.

%% file: sections/experiments.tex
\begin{wrapfigure}{R}{0.55\textwidth}

    \centering{
    \vspace{-1.2em}\includegraphics[width=0.55\columnwidth]{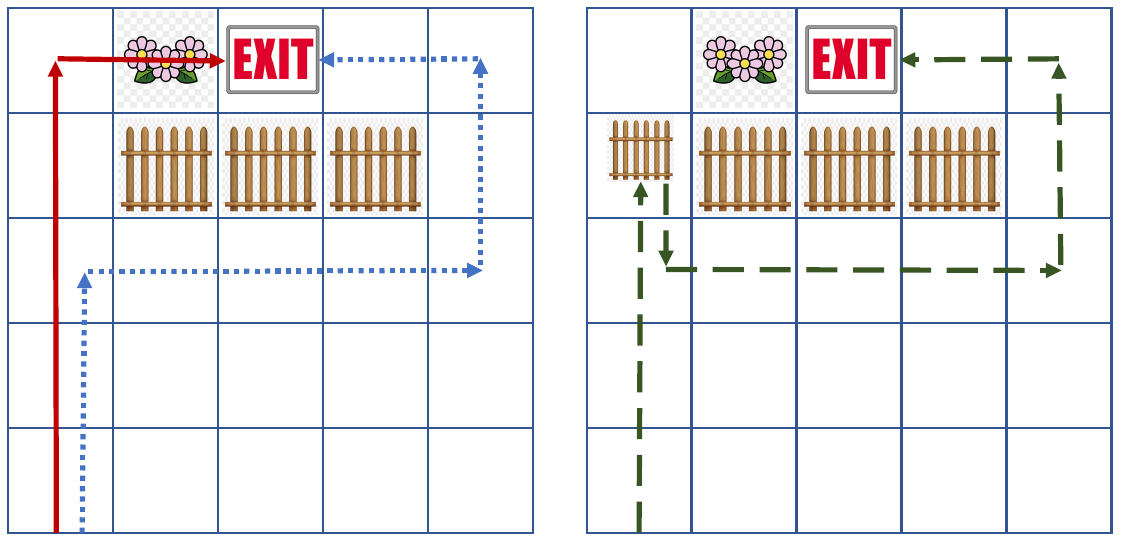}}
    \vspace*{-0.5em}
   \caption{Different caring coefficients lead to different paths. Inconsiderate agents trample the flowers. Super considerate agents build a fence to protect the flowers.}
    \label{fig:diff}

\vspace{-1em}
\end{wrapfigure}

Figure~\ref{fig:diff}, taken from~\citep{alizadeh2022considerate}, illustrates the difference in treatment of individuals through the choice of~\caring~coefficients when using the modified reward in Eq.~\eqref{eq:rvaluei}. There are three agents (our AI agent (agent 1), Alice (agent 2), and Bob (agent 3)) that each want to get to the exit from the starting point at the bottom left. Agents get -1 reward for each step. Alice has a garden and gets very upset (-20 reward) if someone walks on her flowers.  
The AI agent cares about Bob and itself in an equal amount ($\alpha_1 = \alpha_3 = 1$).

In the first case we consider, the AI agent is oblivious to Alice ($\alpha_2 = 0$) and follows the shortest path to the exit, passing through Alice's garden {\textcolor{red}{(red path in the left figure)}}, making her upset. In the second case, the AI agent cares about Alice a little ($\alpha_2 = 1$) and takes the longer path {\textcolor{blue}{(blue path in the left figure)}} to avoid passing through the garden. In the third case, the AI agent cares about Alice a lot ($\alpha_2 = 10$) and even though there is a reward of -50, the AI agent builds a fence to protect Alice's garden {\textcolor{ForestGreen}{(green path in the right figure)}}, which also makes Bob take the longer path with extra steps.

%% file: sections/options.tex
\subsection{Using Information about Options}
\label{sec:options}
In Section \ref{sec:introduction} we posited that a pluralistically-aligned agent should endeavour to act in a manner that ensures the agency of other agents, in service of the social welfare of the collective. Rather than use a distribution over value functions as a proxy for agents' agency, here we consider other agents to instead be endowed with a set of \emph{options} \citep{Sutton1999options} that could reflect particular skills or tasks they are capable of realizing, and we use a distribution over such options to characterize the extent of their agency.

An option is a tuple $\tuple{\initset,\pi,\beta}$ where $\initset\subseteq S$ is the initiation set, $\pi$ is a policy, and $\beta$ is a termination condition (formally, a function associating each state with a termination probability) \citep{Sutton1999options}. The idea is that an agent can follow an option by starting from a state in its initiation set $\initset$ and following the policy $\pi$ until it terminates. Options provide a form of macro action that can be used as a temporally abstracted building block in the construction of policies. Options are often used in
Hierarchical RL: an agent can learn a policy to choose options to execute instead of actions. Here we will use options to represent skills or tasks that other agents in the environment may wish to perform.

Suppose we have a set $\mathcal{O}$ of initiation sets of options, and a probability function $\Pnext(\initset)$ giving the probability that $\initset$ is the initiation set of the option whose execution will be attempted after the AI agent reaches a terminating state.
To try to make the acting agent act so as to allow the execution of that option, we can modify the AI agent's reward function $r_1$, yielding the new reward function $\rnexto$ below.
\begin{align}
\label{eq:roption}
\begin{aligned}
    & \rnexto(s,a,s')= \begin{cases}\alpha_1\cdot r_1(s,a,s')&\hspace*{-.24in}\text{if $s'$ is not terminal}\\
    \alpha_1\cdot r_1(s,a,s')+\gamma\cdot \alpha_2\textstyle\sum_{\initset\in\mathcal{O}} \Pnext(\initset)\cdot \mathds{I}_\initset(s') &\hspace*{-.04in}\text{if $s'$ is terminal}
    \end{cases}
\end{aligned}
\end{align}
where $\mathds{I}_{\initset}:S\to \{0,1\}$ is the indicator function for $\initset$ as a subset of $S$, i.e., $\mathds{I}_{\initset}(s)=\begin{cases}1&\text{if }s\in\initset\\0&\text{otherwise}\end{cases}$.

Note that if $\mathcal{O}$ is finite and $\Pnext$ is a uniform distribution, then the auxiliary reward given by $\rnexto$ will be proportional to how many options in $\mathcal{O}$ can be started in the terminal state.

The hyperparameters $\alpha_1$ and $\alpha_2$ determine how much weight is given to the original reward function and to the ability %of the second agent to initiate its option.
to initiate the option. Given a fixed value of $\alpha_1$ (and ignoring the discount factor), the parameter $\alpha_2$ could be understood as a ``budget'', indicating how much negative reward the acting agent is willing to endure in order to let the option get executed.

Finally, if we had a distribution over pairs $\tuple{\initset,V}$ -- consisting of an option's initiation set and a value function associated with that -- then 
this can be captured by the following augmentation:
\begin{align}
   \begin{aligned}
    & \rnexto'(s,a,s')= \begin{cases}
    \alpha_1\cdot r_1(s,a,s')\hspace*{2.22in} \text{if $s'$ is not terminal} \\
    \alpha_1\cdot r_1(s,a,s')  + \alpha_2%\displaystyle
    \sum_{\tuple{\initset,V}\in\mathcal{O}}  \Pnext(\tuple{\initset,V})\cdot \mathds{I}_\initset(s')\cdot V(s')%\\
     %\hspace*{2.8in}
     \text{ if $s'$ is terminal} 
    \end{cases}
    \end{aligned}
\end{align}
This is much like $\rnexto$ but has an extra factor of $V(s')$ in the sum in the second case.

%% file: sections/concluding-remarks.tex
\section{Concluding Remarks}
\label{sec:concrem}

In this paper we reflected on the issue of pluralistic alignment in the context of agentic AI. We posited (here, in brief) that a pluralistically well-aligned agent should attempt to realize its goals in a manner that contributes towards maximizing the social welfare of the collective, and ensuring the agency of other agents in service of that welfare. As an example of how we might achieve this, we wanted to share a selection of results from a 2022 paper by the authors that constructs an RL agent to learn a policy that is pluralistically aligned in so far as it learns a policy that optimizes for the wellbeing and future agency of a collection of agents, employing a distribution over value functions to serve as a proxy for their agency. That paper \citep{alizadeh2022considerate} provides further details regarding the approach as well as a discussion of related work. As we noted at the outset, defining pluralistic alignment for agentic systems is complex and while this work does not resolve that challenge, we believe it presents a viewpoint and some possible first steps towards realizing agents that are pluralistically aligned.

%% file: sections/acknowledgements.tex
\section*{Acknowledgements}

We gratefully acknowledge funding from the Natural Sciences and
Engineering Research Council of Canada (NSERC) and the Canada CIFAR AI Chairs
Program. The second author also received funding from Open Philanthropy. The third author also acknowledges funding from the National Center for Artificial Intelligence CENIA FB210017 (Basal ANID) and Fondecyt grant 11230762. Resources used in preparing this research
were provided, in part, by the Province of Ontario, the Government of
Canada through CIFAR, and companies sponsoring the Vector Institute for
Artificial Intelligence (\url{https://vectorinstitute.ai/partnerships/}). Finally, we
thank the Schwartz Reisman Institute for Technology and Society for
providing a rich multi-disciplinary research environment.

%% file: main.bbl
\begin{thebibliography}{8}
\providecommand{\natexlab}[1]{#1}
\providecommand{\url}[1]{\texttt{#1}}
\expandafter\ifx\csname urlstyle\endcsname\relax
  \providecommand{\doi}[1]{doi: #1}\else
  \providecommand{\doi}{doi: \begingroup \urlstyle{rm}\Url}\fi

\bibitem[Alamdari et~al.(2022)Alamdari, Klassen, Toro~Icarte, and McIlraith]{alizadeh2022considerate}
Parand~A. Alamdari, Toryn~Q. Klassen, Rodrigo Toro~Icarte, and Sheila~A. McIlraith.
\newblock Be considerate: Avoiding negative side effects in reinforcement learning.
\newblock In \emph{Proceedings of the 21st International Conference on Autonomous Agents and Multiagent Systems}, pages 18--26. International Foundation for Autonomous Agents and Multiagent Systems (IFAAMAS), 2022.

\bibitem[Alamdari et~al.(2024)Alamdari, Ebadian, and Procaccia]{policyaggregation}
Parand~A. Alamdari, Soroush Ebadian, and Ariel~D. Procaccia.
\newblock Policy aggregation.
\newblock In \emph{Proceedings of the 38th Annual Conference on Neural Information Processing Systems (NeurIPS)}, 2024.

\bibitem[Sen(1974)]{Sen1974rawls}
Amartya Sen.
\newblock Rawls versus {B}entham: An axiomatic examination of the pure distribution problem.
\newblock \emph{Theory and Decision}, 4\penalty0 (3-4):\penalty0 301--309, 1974.
\newblock \doi{10.1007/BF00136651}.

\bibitem[Shavit et~al.(2023)Shavit, Agarwal, Brundage, Adler, O’Keefe, Campbell, Lee, Mishkin, Eloundou, Hickey, Slama, Ahmad, McMillan, Beutel, Passos, and Robinson]{shavit2023practices}
Yonadav Shavit, Sandhini Agarwal, Miles Brundage, Steven Adler, Cullen O’Keefe, Rosie Campbell, Teddy Lee, Pamela Mishkin, Tyna Eloundou, Alan Hickey, Katarina Slama, Lama Ahmad, Paul McMillan, Alex Beutel, Alexandre Passos, and David~G. Robinson.
\newblock Practices for governing agentic {AI} systems.
\newblock Research paper, OpenAI, December 2023.

\bibitem[Sorensen et~al.(2024)Sorensen, Moore, Fisher, Gordon, Mireshghallah, Rytting, Ye, Jiang, Lu, Dziri, Althoff, and Choi]{SorensenICML2024roadmap}
Taylor Sorensen, Jared Moore, Jillian Fisher, Mitchell~L Gordon, Niloofar Mireshghallah, Christopher~Michael Rytting, Andre Ye, Liwei Jiang, Ximing Lu, Nouha Dziri, Tim Althoff, and Yejin Choi.
\newblock Position: A roadmap to pluralistic alignment.
\newblock In \emph{Proceedings of the 41st International Conference on Machine Learning}, pages 46280--46302. PMLR, 21--27 Jul 2024.

\bibitem[Sutton and Barto(2018)]{Sutton2018textbook}
Richard~S. Sutton and Andrew~G. Barto.
\newblock \emph{Reinforcement Learning: An Introduction}.
\newblock MIT Press, Cambridge, MA, second edition, 2018.

\bibitem[Sutton et~al.(1999)Sutton, Precup, and Singh]{Sutton1999options}
Richard~S. Sutton, Doina Precup, and Satinder~P. Singh.
\newblock Between {MDP}s and semi-{MDP}s: {A} framework for temporal abstraction in reinforcement learning.
\newblock \emph{Artificial Intelligence}, 112\penalty0 (1-2):\penalty0 181--211, 1999.
\newblock \doi{10.1016/S0004-3702(99)00052-1}.

\bibitem[Weymark(1981)]{weymark1981generalized}
John~A Weymark.
\newblock Generalized {G}ini inequality indices.
\newblock \emph{Mathematical Social Sciences}, 1\penalty0 (4):\penalty0 409--430, 1981.

\end{thebibliography}
